\RequirePackage{silence}
\documentclass[runningheads]{llncs}

\usepackage[T1]{fontenc}
\usepackage{hyperref}
\usepackage{float}
\usepackage{graphicx}
\usepackage{caption}
\usepackage{subcaption}
\usepackage[table]{xcolor}
\usepackage{tabularx}
\usepackage{url}
\usepackage{booktabs}
\usepackage{makecell}
\usepackage{array}
\usepackage{multirow}
\usepackage{amsmath,amssymb}
\usepackage{enumitem}
\usepackage{array}

\begin{document}

\title{Crop Yield Prediction Using Regression Models: A Comparative Evaluation of Model Performance}
\titlerunning{Crop Yield Prediction Using Regression Models}

\author{Saptak Das\inst{1} \and
Arijit Sarkar\inst{2} \and
Soumik Neogi\inst{2} \and
Suborno Naskar\inst{2}}

\authorrunning{S. Das et al.}

\institute{Indian Institute of Science Education and Research
Thiruvananthapuram, Kerala, India
\and
Department of Information Technology, MCKV Institute of Engineering,
Howrah, West Bengal, India}

\maketitle

\begin{abstract}
Reliable crop-yield prediction can support agricultural planning and resource allocation. This study evaluates five regression models for predicting crop yield from environmental, crop-related, and agricultural management variables. The dataset initially contained 1,000,000 observations and ten variables, with 231 observations having negative target values removed during preprocessing. Categorical variables were represented using one-hot encoding, and Boolean variables were converted into binary values. The models considered were multiple linear regression (LR), random-forest regression (RF), decision-tree regression (DT), ridge regression, and lasso regression. Their performance was evaluated using 80:20 and 70:30 training--testing splits based on the coefficient of determination ($R^2$), mean squared error (MSE), and root mean squared error (RMSE). Applying an $80:20$ split, LR achieved a test $R^2$ of approximately $0.9130$ and a test RMSE of $0.5008$. Applying a $70:30$ split, lasso regression yielded the highest test $R^2$ (approx. $0.9131$) and the lowest test RMSE (approx. $0.5008$), with performance almost identical to LR and ridge regression. DT also showed improved test performance under the 70:30 split, whereas RF exhibited a substantial gap between training and testing performance. These results indicate that the evaluated linear regression model provides competitive baselines for this dataset. Further experimentation, including validation testing, will help determine the models' performance and predictability.

\end{abstract}

\section{Introduction}

Crop yield is influenced by environmental, biological, and agricultural management factors. Accurate modeling of expected crop yields assists crop and field management, including planning and resource allocation. Predicting crop yields is a rapidly advancing area of agricultural science and machine learning, owing in part to the pervasiveness of crop diseases and the negative effect of diseases on crop yields \cite{liakos2018,vanKloppenburg2020}. However, the performance of prediction models is highly data-dependent and influenced by the training and evaluation strategies.

This study investigates crop-yield prediction using a dataset containing regional, soil, crop, weather, and management information. The predictor variables include region, soil type, crop type, rainfall, temperature, fertilizer use, irrigation use, weather condition, and days to harvest. The target variable, \texttt{Yield\_tons\_per\_hectare}, represents crop yield in tonnes per hectare. During preprocessing, observations with negative target values were removed because they are not physically meaningful as crop yields in this study.

This dataset will require a decision on how to encode categorical variables. One-hot encoding converts each category into a binary indicator. Using this method results in 24 features. However, by removing one indicator for each of the four categorical variables, the feature representation can be reduced to 20. Removing the indicators results in a similar impact on the linear dependence among the indicator variables as including the reference category. Therefore, the way we represent the features affects the model specification and interpretation of the coefficients. This may have an effect on the predictive power of the model.

We look at five different regression methods: multiple linear regression, random-forest regression, decision-tree regression, ridge regression, and lasso regression. This allows us to evaluate a traditional linear approach with tree-based and regularized linear approaches. We evaluated our models with both an 80:20 and a 70:30 training-testing split. We used GridSearchCV and RandomizedSearchCV separately to tune the hyperparameters. We also evaluated models without any tuning. The performance of the models was evaluated with $R^2$, MSE, and RMSE, on both the training and testing sets.

The results demonstrate that, for both split ratios, LR, ridge regression and lasso regression achieved test $R^2$ values of approximately 0.9130. Lasso regression achieved the lowest test MSE and RMSE for the 70:30 split. DT achieved a higher test $R^2$ for the 70:30 split than for the 80:20 split. For RF, the training and test $R^2$ values differed widely, indicating that, in this case, the training performance was not indicative of the test performance. Differences among some experimental settings (e.g., hyperparameter tuning, feature selection, etc.) lead to the following conclusions being drawn about the evaluated configurations rather than definitive statements regarding algorithm and split performance.

\vspace{0.3cm}

The main contributions of this study are:
\begin{enumerate}[leftmargin=*]
    \item Preprocessing and representation of crop-yield data containing environmental, categorical, and agricultural management variables;
    \item Evaluation of five regression models under 80:20 and 70:30 training--testing splits using $R^2$, MSE, and RMSE;
    \item Comparative analysis of the observed training and testing performance, highlighting similarities among linear-model approaches and the generalisation gap observed for random-forest regression.
\end{enumerate}

\section{Dataset Description}
The dataset contains 1,000,000 rows and 10 variables. The dataset used in this study was obtained from Kaggle:
\textit{Agriculture Crop Yield} by Samuel Oti Attakorah~\cite{kagglecropyield}.
 The original dataset was used as the basis for the preprocessing and regression experiments described in this study. During data preprocessing, we identified 231 observations with negative values for the target variable, \textit{Yield\_tons\_per\_hectare}. Since negative crop yield values are not physically meaningful in the context of this study, these observations were excluded from the dataset. Consequently, the remaining 999,769 observations were retained for subsequent analysis and model development. The variables and their roles are summarised in Table~\ref{tab:variables}.

\begin{table}[H]
\centering
\caption{Variables in the supplied raw dataset.}
\label{tab:variables}
\small
\setlength{\tabcolsep}{4pt}
\renewcommand{\arraystretch}{1.3}

\begin{tabular}{|
>{\centering\arraybackslash}m{0.30\textwidth}|
>{\centering\arraybackslash}m{0.17\textwidth}|
>{\raggedright\arraybackslash}m{0.41\textwidth}|}
\hline\hline
\centering\arraybackslash\textbf{Variable} &
\centering\arraybackslash\textbf{Role} &
\centering\arraybackslash\textbf{Description} \\
\hline \hline 
Region
& Predictor
& Geographic region (East, North, South, West) \\
\hline
Soil\_Type
& Predictor
& Soil category (Chalky, Clay, Loam, Peaty, Sandy, Silt) \\
\hline
Crop
& Predictor
& Crop category (Barley, Cotton, Maize, Rice, Soybean, Wheat) \\
\hline
Rainfall\_mm
& Predictor
& Rainfall in millimetres \\
\hline
Temperature\_Celsius
& Predictor
& Temperature in degrees Celsius \\
\hline
Fertilizer\_Used
& Predictor
& Boolean indicator of fertilizer use \\
\hline
Irrigation\_Used
& Predictor
& Boolean indicator of irrigation use \\
\hline
Weather\_Condition
& Predictor
& Weather category (Cloudy, Rainy, Sunny) \\
\hline
Days\_to\_Harvest
& Predictor
& Number of days to harvest \\
\hline
Yield\_tons\_per\_hectare
& Target
& Crop yield in tonnes per hectare \\
\hline
\end{tabular}
\end{table}

The dataset contains no missing values in any of its ten columns. It includes four region categories, six soil categories, six crop categories, and three weather categories. Rainfall ranges from approximately 100 to $1,000$ mm, temperature from $15$ to $40$ degrees Celsius, and days to harvest from $60$ to $149$ days. The raw dataset contains $231$ observations with negative target values, which were removed during preprocessing. After their removal, the target variable, Yield\_tons\_per\_hectare, ranges from approximately $0.00041$ to $9.963$ tonnes per hectare across the remaining $999,769$ observations.

\section{Related Work}
Machine learning techniques have been widely reviewed for agricultural applications, including crop and livestock management, disease detection, and yield estimation \cite{liakos2018}. A systematic review by van Klompenburg et al. discusses machine-learning methods used for crop-yield prediction and the factors affecting reported performance \cite{vanKloppenburg2020}. These reviews motivate careful consideration of data quality, predictor construction, validation strategy, and the generalisability of reported scores.

The algorithms considered here are established methods for supervised regression. Decision trees learn a sequence of feature-based partitions \cite{quinlan1986}; random forests aggregate predictions from many trees to reduce variance and improve robustness \cite{breiman2001}. Linear regression provides an interpretable linear baseline. Ridge regression penalises the squared magnitude of coefficients and can stabilise estimates in the presence of correlated predictors \cite{hoerl1970}, while lasso regression uses an absolute-value penalty that can shrink some coefficients to zero \cite{tibshirani1996}. These models give comparative insights between linear and regularized linear models as well as tree based models.

\section{Methodology}
\subsection{Data preprocessing}
The preprocessing was performed in the following sequence:
\begin{enumerate}[leftmargin=*]
    \item \textbf{Check for missing values:} No values were found to be missing.
    \item \textbf{Remove negative values of the target variable:} The dataset contained $1,000,000$ rows. $231$ observations were removed as their Yield\_tons\_per\_hectare values were negative, leaving $999,769$ observations for further analysis.
    \item \textbf{One hot encoding:} \texttt{Region}, \texttt{Soil\_Type}, \texttt{Crop}, and \texttt{Weather\_Condition} were converted into binary indicators. 
    \item \textbf{Boolean-to-binary conversion:} \texttt{Fertilizer\_Used} and \texttt{Irrigation\_Used} were represented as 1 for \texttt{True} and 0 for \texttt{False}. This is a binary recoding of Boolean variables, not one-hot encoding.
    \item \textbf{Feature designs:} The full one-hot design retained all 19 categorical indicator columns plus five other predictors (rainfall, temperature, fertilizer use, irrigation use, and days to harvest), giving 24 predictors.The reference-dropped design removed one of the indicators from each of the four categorical groups, leaving 20 predictors.
    \item \textbf{Target separation:} \texttt{Yield\_tons\_per\_hectare} was taken as the continuous target and was not included in the set of predictors.
\end{enumerate}

The 24 feature model consists of 4 regional indicators, 6 soil indicators, 6 crop indicators, 3 weather indicators and 5 additional indicators. In the 20-feature design, one category from each group is used as the reference level. For ordinary least squares with an intercept, this avoids exact linear dependence among each set of one-hot indicators. While the complete representation might be of value for tree based models and regularized regression, interpretation of coefficients for the complete one-hot design should account for redundancy and treatment of the model with respect to the intercept and regularization.

\subsection{Machine-learning models}
The five regression methods considered were:
\begin{itemize}[leftmargin=*]
    \item \textbf{Multiple linear regression (LR):} Estimates a linear relationship between the continuous target and a set of predictors
    \item \textbf{Random forest regression (RF):} Uses the average of predictions of an ensemble of decision trees \cite{breiman2001}.
    \item \textbf{Decision-tree regression (DT):} Partitions the feature space into regions and predicts a value for each terminal region \cite{quinlan1986}.
    \item \textbf{Ridge regression:} Minimises squared error with $L_2$ coefficient penalty \cite{hoerl1970}.
    \item \textbf{Lasso regression:} Minimises squared error with $L_1$ coefficient penalty \cite{tibshirani1996}.
\end{itemize}

\subsection{Training and hyperparameter-search protocol}
Two train–test ratios were considered: 80:20 and 70:30. For each feature design, experiments were conducted both without hyperparameter tuning and with hyperparameter tuning using RandomizedSearchCV or GridSearchCV. The configuration demonstrating better predictive performance was retained for reporting. Hyperparameter search was conducted using the training data only, with the test set held out for final evaluation. The best hyperparameters were recorded for each model where applicable. The reported results therefore represent the selected configurations rather than a uniform comparison of tuned and untuned models.

\subsection{Evaluation metrics}
For each algorithm $R^2$, mean squared error, and root mean squared error were calculated. 

The coefficient of determination is
\[
R^2 = 1-\frac{\sum_{i=1}^{n}(y_i-\hat{y}_i)^2}
{\sum_{i=1}^{n}(y_i-\bar{y})^2},
\]
where $y_i$ and $\hat{y}_i$ are the observed and predicted yields respectively and $\bar{y}$ is the observed mean. The mean squared error and root mean squared error are
\[
\mathrm{MSE}=\frac{1}{n}\sum_{i=1}^{n}(y_i-\hat{y}_i)^2,
\qquad
\mathrm{RMSE}=\sqrt{\mathrm{MSE}}.
\]
Higher test $R^2$ and lower test MSE/RMSE indicate better performance, provided the models are evaluated on the same test observations. 

\section{Experimental Study}
\subsection{Experimental Configurations}

The experiments evaluated five regression models: multiple linear
regression (LR), random-forest regression (RF), decision-tree
regression (DT), ridge regression, and lasso regression. Two
training--testing ratios, namely 80:20 and 70:30, were considered.
Model performance was evaluated using $R^2$, mean squared error, and root mean
squared error. Training metrics were used to assess the
fit on the training data, whereas testing metrics were used to
compare predictive performance on unseen observations.

The experimental configurations were not identical across all
models. For the 80:20 split, LR used 20 predictors without
hyperparameter tuning, RF used 24 predictors without tuning,
DT used 24 predictors with RandomizedSearchCV, ridge regression
used 24 predictors without tuning, and lasso regression used
24 predictors with GridSearchCV.  In the 70:30 split, the five models used 24 predictors, of which, for Lasso regression and Decision Trees, Grid Search was used to tune and for Linear regression, Ridge regression and Random forest regression, no hyperparameter tuning was done. The configurations are summarised
in Table~\ref{tab:experimental-config}.

\begin{table}[H]
\centering
\caption{Experimental configurations for the two train--test splits.}
\label{tab:experimental-config}
\small
\setlength{\tabcolsep}{4pt}
\renewcommand{\arraystretch}{1.3}

\begin{tabular}{|
>{\centering\arraybackslash}m{0.16\textwidth}|
>{\centering\arraybackslash}m{0.36\textwidth}|
>{\centering\arraybackslash}m{0.36\textwidth}|}
\hline\hline
\textbf{Model} &
\makecell[c]{\textbf{80:20 }\textbf{split}} &
\makecell[c]{\textbf{70:30 }\textbf{split}} \\
\hline \hline 
LR
& 20 predictors; untuned
& 24 predictors; untuned \\
\hline
RF
& 24 predictors; untuned
& 24 predictors; untuned \\
\hline
DT
& 24 predictors; random search
& 24 predictors; grid search \\
\hline
Ridge
& 24 predictors; untuned
& 24 predictors; untuned \\
\hline
Lasso
& 24 predictors; grid search
& 24 predictors; grid search \\
\hline
\end{tabular}
\end{table}

\subsection{Results for the 80:20 Train--Test Split}

The training and testing results for the 80:20 split are presented in Table~\ref{tab:results-8020}. LR, Ridge Regression, and Lasso Regression achieved very similar test $R^2$ (approximately 0.9130), with the test MSE (approximately 0.250777) and RMSE (approximately 0.500776) of the testing set. Test performance of ridge and lasso regression was almost the same. The minor differences among the three models indicate that their predictive performance was similar in the investigated configurations.

RF had the highest training  $R^2$ (approximately 0.9871) and lowest training errors. However, its test $R^2$
decreased to approximately 0.9077, and its test RMSE increased
to approximately 0.5157. This training--testing gap suggests
that RF fitted the training observations more closely than
the other models but generalised less effectively on the
held-out test set. DT achieved a test $R^2$ of approximately
0.9100, placing its performance between RF and the three
linear-model variants.

% Table 3: Regression performance for the 80:20 train-test split

\begin{table}[H]
\centering
\caption{Regression performance for the 80:20 train--test split.}
\label{tab:results-8020}
\scriptsize
\setlength{\tabcolsep}{3pt}
\renewcommand{\arraystretch}{1.5}

\resizebox{\textwidth}{!}{%
\begin{tabular}{|
>{\centering\arraybackslash}m{1.7cm}|
r|r|r|r|r|r|}
\hline
\hline
\multicolumn{1}{|>{\centering\arraybackslash}m{1.7cm}|}{\textbf{Model}}
& \multicolumn{1}{>{\centering\arraybackslash}m{1.25cm}|}{\textbf{Tr. $R^2$}}
& \multicolumn{1}{>{\centering\arraybackslash}m{1.25cm}|}{\textbf{Tr. MSE}}
& \multicolumn{1}{>{\centering\arraybackslash}m{1.25cm}|}{\textbf{Tr. RMSE}}
& \multicolumn{1}{>{\centering\arraybackslash}m{1.25cm}|}{\textbf{Te. $R^2$}}
& \multicolumn{1}{>{\centering\arraybackslash}m{1.25cm}|}{\textbf{Te. MSE}}
& \multicolumn{1}{>{\centering\arraybackslash}m{1.25cm}|}{\textbf{Te. RMSE}}
\\
\hline
\hline
LR
& 0.9129706418 & 0.2504014163 & 0.5004012553
& 0.9130137709 & 0.2507766686 & 0.5007760663 \\
\hline
RF
& 0.9870627444 & 0.03722315308 & 0.1929330275
& 0.9077351184 & 0.2659947428 & 0.5157467817 \\
\hline
DT
& 0.9161022839 & 0.2413910359 & 0.4913156174
& 0.909994562 & 0.2594808871 & 0.5093926649 \\
\hline
Ridge Regression
& 0.9129706417 & 0.2504014163 & 0.5004012553
& 0.9130137621 & 0.250776694 & 0.5007760917 \\
\hline
Lasso Regression
& 0.9129703752 & 0.2504021831 & 0.5004020215
& 0.9130138109 & 0.2507765533 & 0.5007759512 \\
\hline
\end{tabular}%
}
\end{table}

\begin{table}[H]
\centering
\caption{Model configurations and hyperparameters for the 80:20 train--test split}
\label{tab:methodology_80_20}
\small
\setlength{\tabcolsep}{4pt}
\renewcommand{\arraystretch}{1.3}

\begin{tabular}{|>{\centering\arraybackslash}m{0.19\textwidth}|
                  >{\centering\arraybackslash}m{0.37\textwidth}|
                  >{\centering\arraybackslash}m{0.30\textwidth}|}
\hline\hline
\textbf{Model} &
\makecell[c]{\textbf{Methodology}\\\textbf{(80:20)}} &
\makecell[c]{\textbf{Hyperparameter}\\\textbf{(If any)}} \\
\hline \hline
LR &
20 features; without hyperparameter tuning &
--- \\
\hline
RF &
24 features; without hyperparameter tuning &
--- \\
\hline
DT &
24 features; with hyperparameter tuning; RandomizedSearchCV &
\makecell[c]{max\_depth = 52\\
min\_samples\_split = 16\\
min\_samples\_leaf = 93} \\
\hline
Ridge Regression &
24 features; without hyperparameter tuning &
--- \\
\hline
Lasso Regression &
24 features; with hyperparameter tuning; GridSearchCV &
alpha = 0.0001 \\
\hline
\end{tabular}
\end{table}

The results for the selected 80:20 train–test configuration are presented in Table~\ref{tab:results-8020}. Detailed results for alternative feature sets and hyperparameter-tuning strategies are provided in the Appendix (Tables~7–12).

\subsection{Results for the 70:30 Train--Test Split}

The results for the 70:30 split are presented in
Table~\ref{tab:results-7030}. LR, ridge regression, and lasso
regression again exhibited very similar test performance,
with test $R^2$ values close to 0.9131. Lasso regression
recorded the lowest test MSE (approximately
0.250770), while LR recorded a test $R^2$ of approximately
0.913092. The differences between these linear-model
variants were small.

DT achieved a test $R^2$ of approximately 0.9120 and a test
RMSE of approximately 0.5040, improving upon its test
performance under the 80:20 split. RF continued to show a
substantial difference between training and testing
performance: its training $R^2$ was approximately 0.9870,
whereas its test $R^2$ was approximately 0.9077. This
pattern is consistent with a possible generalisation gap.

\begin{table}[H]
\centering
\caption{Regression performance for the 70:30 train--test split.}
\label{tab:results-7030}
\scriptsize
\setlength{\tabcolsep}{3pt}
\renewcommand{\arraystretch}{1.5}

\resizebox{\textwidth}{!}{%
\begin{tabular}{|
>{\centering\arraybackslash}m{1.7cm}|
r|r|r|r|r|r|}
\hline
\hline
\multicolumn{1}{|>{\centering\arraybackslash}m{1.7cm}|}{\textbf{Model}}
& \multicolumn{1}{>{\centering\arraybackslash}m{1.25cm}|}{\textbf{Tr. $R^2$}}
& \multicolumn{1}{>{\centering\arraybackslash}m{1.25cm}|}{\textbf{Tr. MSE}}
& \multicolumn{1}{>{\centering\arraybackslash}m{1.25cm}|}{\textbf{Tr. RMSE}}
& \multicolumn{1}{>{\centering\arraybackslash}m{1.25cm}|}{\textbf{Te. $R^2$}}
& \multicolumn{1}{>{\centering\arraybackslash}m{1.25cm}|}{\textbf{Te. MSE}}
& \multicolumn{1}{>{\centering\arraybackslash}m{1.25cm}|}{\textbf{Te. RMSE}}
\\
\hline
\hline
LR
& 0.9129304560 & 0.2503510489 & 0.5003509258
& 0.9130916682 & 0.2507726187 & 0.5007720227 \\
\hline
RF
& 0.9870488887 & 0.0372383284 & 0.1929723516
& 0.9077328853 & 0.2662353023 & 0.5159799437 \\
\hline
DT
& 0.9134832049 & 0.2487617300 & 0.4987601929
& 0.9119788539 & 0.2539836270 & 0.5039678829 \\
\hline
Ridge Regression
& 0.912930456 & 0.25035104897 & 0.5003509258
& 0.913091663 & 0.25077263296 & 0.500772036 \\
\hline
Lasso Regression
& 0.9129301770 & 0.2503518512 & 0.500351727
& 0.9130925912 & 0.2507699556 & 0.500769363 \\
\hline
\end{tabular}%
}
\end{table}

\begin{table}[H]
\centering
\caption{Model configurations and hyperparameters for the 70:30 train--test split}
\label{tab:methodology_70_30}
\small
\setlength{\tabcolsep}{4pt}
\renewcommand{\arraystretch}{1.3}

\begin{tabular}{|
>{\centering\arraybackslash}m{0.19\textwidth}|
>{\centering\arraybackslash}m{0.37\textwidth}|
>{\centering\arraybackslash}m{0.30\textwidth}|}
\hline\hline
\textbf{Model} &
\makecell[c]{\textbf{Methodology}\\\textbf{(70:30)}} &
\makecell[c]{\textbf{Hyperparameter}\\\textbf{(If any)}} \\
\hline \hline
LR &
24 features; without hyperparameter tuning &
--- \\
\hline
RF &
24 features; without hyperparameter tuning &
--- \\
\hline
DT &
24 features; with hyperparameter tuning; GridSearchCV &
\makecell[c]{max\_depth = 10\\
min\_samples\_split = 2\\
min\_samples\_leaf = 4} \\
\hline
Ridge Regression &
24 features; without hyperparameter tuning &
--- \\
\hline
Lasso Regression &
24 features; with hyperparameter tuning; GridSearchCV &
alpha = 0.0001 \\
\hline
\end{tabular}
\end{table}

The results for the selected 70:30 train–test configuration are presented in Table~\ref{tab:results-7030}. Detailed results for alternative feature sets and hyperparameter-tuning strategies are provided in the Appendix (Tables~13–18).

\subsection{Comparative Discussion}

Across the two train--test splits, the test $R^2$ values of
LR, ridge regression, and lasso regression remained close to
0.913. Their test RMSE values were approximately 0.501,
indicating similar prediction errors on the scale of the
target variable. Lasso regression yielded the lowest displayed
test MSE and RMSE for the 70:30 experiment, although its
advantage over LR and ridge regression was marginal.

In contrast, RF achieved substantially better training scores
than testing scores, suggesting that its fitted predictions
may be more sensitive to the training observations. DT showed
a smaller training--testing gap under the 70:30 split than
under the 80:20 split and obtained a higher test $R^2$ in the
70:30 experiment.

For completeness, the detailed regression performance results for all evaluated experimental configurations are provided in the Appendix. These results cover the 80:20 and 70:30 train–test splits, the use of 20 and 24 features, and the three hyperparameter-tuning settings: no hyperparameter tuning, RandomizedSearchCV, and GridSearchCV. The Appendix enables comparison of the evaluated models across these configurations, while the main text focuses on the selected experimental results.

\section{Conclusion and Future Scope}
This research examined five regression models of crop-yield prediction using environmental, crop and agricultural management variables. The experiments considered 80:20 and
70:30 train--test splits and evaluated the models using
$R^2$, MSE, and RMSE. It was found that LR, ridge
regression, and lasso regression showed similar test
performance, with $R^2$ values close to 0.913 and RMSE values
of approximately 0.501. Lasso regression produced the lowest
reported test MSE and RMSE in the 70:30 experiment, although
the differences from LR and ridge regression were small.
DT showed competitive results, especially under the 70:30 split. RF, in general, showed high training results but comparatively low test results, thus showing a case of over-fitting.

The results indicate that the linear model provides a strong baseline for the dataset under consideration. However, the experiments used various configurations for some models and, therefore, the results can be interpreted as comparisons of the configurations rather than evidence to support the overall superiority of the algorithm or train-test ratio. Moreover, results show predictive performance over the dataset.

In future work, experiments will be designed to study the impact of diverse configurations, controlled data partitions and systematic hyperparameter tuning on all the models involved. The stability of the results can be evaluated by using a repeated train-test split or cross-validation of the training data. Other methods of regression, for example, boosting methods, can be assessed. Also, methods to determine feature importance can help identify the impact of variables such as rainfall, soil type and crop type, as well as irrigation, fertilizer use and harvesting, on predicted yield. Assessing the models on independently collected data would help determine how well the models generalize. The models would need to be assessed on data collected during varying growing seasons and in different locales. The models, ultimately, should help inform agricultural decisions.

\section{Additional Experimental Results (Appendix)}
\label{app:additional-results}

This appendix presents the results of the alternative experimental configurations evaluated in this study, covering the 80:20 and 70:30 train–test splits, the use of 20 and 24 features, and models evaluated with and without hyperparameter tuning using RandomizedSearchCV and GridSearchCV

\begin{table}[H]
\centering
\caption{Regression performance for the 80:20 train--test split without hyperparameter tuning using 20 features.}
\label{tab:results-8020-untuned-20}
\scriptsize
\setlength{\tabcolsep}{3pt}
\renewcommand{\arraystretch}{1.5}

\resizebox{\textwidth}{!}{%
\begin{tabular}{|
>{\centering\arraybackslash}m{1.7cm}|
r|r|r|r|r|r|}
\hline
\hline
\textbf{Model}
& \multicolumn{1}{>{\centering\arraybackslash}m{1.25cm}|}{\textbf{Tr. $R^2$}}
& \multicolumn{1}{>{\centering\arraybackslash}m{1.25cm}|}{\textbf{Tr. MSE}}
& \multicolumn{1}{>{\centering\arraybackslash}m{1.25cm}|}{\textbf{Tr. RMSE}}
& \multicolumn{1}{>{\centering\arraybackslash}m{1.25cm}|}{\textbf{Te. $R^2$}}
& \multicolumn{1}{>{\centering\arraybackslash}m{1.25cm}|}{\textbf{Te. MSE}}
& \multicolumn{1}{>{\centering\arraybackslash}m{1.25cm}|}{\textbf{Te. RMSE}}
\\
\hline
\hline
LR
& 0.9129706418 & 0.2504014163 & 0.5004012553
& 0.9130137709 & 0.2507766686 & 0.5007760663 \\
\hline
RF
& 0.987018294 & 0.03735104615 & 0.1932641874
& 0.907479335 & 0.2667321528 & 0.5164611823 \\
\hline
DT
& 1 & 0 & 0
& 0.8156732457 & 0.5314042218 & 0.7289747745 \\
\hline
Ridge Regression
& 0.9129706417 & 0.2504014163 & 0.5004012553
& 0.9130137621 & 0.250776694 & 0.5007760917 \\
\hline
Lasso Regression
& 0.8847181529 & 0.3304832303 & 0.5748767088
& 0.885137457 & 0.3323514293 & 0.5764992882 \\
\hline
\end{tabular}%
}
\end{table}

\begin{table}[H]
\centering
\caption{Regression performance for the 80:20 train--test split using RandomizedSearchCV with 20 features.}
\label{tab:results-8020-random-20}
\scriptsize
\setlength{\tabcolsep}{3pt}
\renewcommand{\arraystretch}{1.5}

\resizebox{\textwidth}{!}{%
\begin{tabular}{|
>{\centering\arraybackslash}m{1.7cm}|
r|r|r|r|r|r|}
\hline
\hline
\textbf{Model}
& \multicolumn{1}{>{\centering\arraybackslash}m{1.25cm}|}{\textbf{Tr. $R^2$}}
& \multicolumn{1}{>{\centering\arraybackslash}m{1.25cm}|}{\textbf{Tr. MSE}}
& \multicolumn{1}{>{\centering\arraybackslash}m{1.25cm}|}{\textbf{Tr. RMSE}}
& \multicolumn{1}{>{\centering\arraybackslash}m{1.25cm}|}{\textbf{Te. $R^2$}}
& \multicolumn{1}{>{\centering\arraybackslash}m{1.25cm}|}{\textbf{Te. MSE}}
& \multicolumn{1}{>{\centering\arraybackslash}m{1.25cm}|}{\textbf{Te. RMSE}}
\\
\hline
\hline
LR
& 0.9129705048 & 0.2504018102 & 0.5004016488
& 0.9130139399 & 0.2507761815 & 0.50077558 \\
\hline
RF
& 0.9441125995 & 0.1607995798 & 0.4009982291
& 0.9092718421 & 0.2615644502 & 0.5114337202 \\
\hline
DT
& 0.9160686443 & 0.2414878239 & 0.4914141063
& 0.9100197862 & 0.2594081671 & 0.5093212808 \\
\hline
Ridge Regression
& 0.9129706415 & 0.250401417 & 0.500401256
& 0.9130137194 & 0.2507768169 & 0.5007762144 \\
\hline
Lasso Regression
& 0.5846677618 & 1.194996525 & 1.093158966
& 0.5836824697 & 1.200221281 & 1.095546111 \\
\hline
\end{tabular}%
}
\end{table}

\begin{table}[H]
\centering
\caption{Regression performance for the 80:20 train--test split using GridSearchCV with 20 features.}
\label{tab:results-8020-grid-20}
\scriptsize
\setlength{\tabcolsep}{3pt}
\renewcommand{\arraystretch}{1.5}

\resizebox{\textwidth}{!}{%
\begin{tabular}{|
>{\centering\arraybackslash}m{1.7cm}|
r|r|r|r|r|r|}
\hline
\hline
\textbf{Model}
& \multicolumn{1}{>{\centering\arraybackslash}m{1.25cm}|}{\textbf{Tr. $R^2$}}
& \multicolumn{1}{>{\centering\arraybackslash}m{1.25cm}|}{\textbf{Tr. MSE}}
& \multicolumn{1}{>{\centering\arraybackslash}m{1.25cm}|}{\textbf{Tr. RMSE}}
& \multicolumn{1}{>{\centering\arraybackslash}m{1.25cm}|}{\textbf{Te. $R^2$}}
& \multicolumn{1}{>{\centering\arraybackslash}m{1.25cm}|}{\textbf{Te. MSE}}
& \multicolumn{1}{>{\centering\arraybackslash}m{1.25cm}|}{\textbf{Te. RMSE}}
\\
\hline
\hline
LR
& 0.9129705048 & 0.2504018102 & 0.5004016488
& 0.9130139399 & 0.2507761815 & 0.50077558 \\
\hline
RF
& 0.9441125995 & 0.1607995798 & 0.4009982291
& 0.9092718421 & 0.2615644502 & 0.5114337202 \\
\hline
DT
& 0.9134173038 & 0.2491162775 & 0.4991154952
& 0.9120229454 & 0.2536331658 & 0.503620061 \\
\hline
Ridge Regression
& 0.9129706417 & 0.2504014163 & 0.5004012553
& 0.9130137621 & 0.250776694 & 0.5007760917 \\
\hline
Lasso Regression
& 0.9129703047 & 0.250402386 & 0.5004022242
& 0.9130134481 & 0.2507775991 & 0.5007769954 \\
\hline
\end{tabular}%
}
\end{table}

\begin{table}[H]
\centering
\caption{Regression performance for the 80:20 train--test split without hyperparameter tuning using 24 features.}
\label{tab:results-8020-untuned-24}
\scriptsize
\setlength{\tabcolsep}{3pt}
\renewcommand{\arraystretch}{1.5}

\resizebox{\textwidth}{!}{%
\begin{tabular}{|
>{\centering\arraybackslash}m{1.7cm}|
r|r|r|r|r|r|}
\hline
\hline
\textbf{Model}
& \multicolumn{1}{>{\centering\arraybackslash}m{1.25cm}|}{\textbf{Tr. $R^2$}}
& \multicolumn{1}{>{\centering\arraybackslash}m{1.25cm}|}{\textbf{Tr. MSE}}
& \multicolumn{1}{>{\centering\arraybackslash}m{1.25cm}|}{\textbf{Tr. RMSE}}
& \multicolumn{1}{>{\centering\arraybackslash}m{1.25cm}|}{\textbf{Te. $R^2$}}
& \multicolumn{1}{>{\centering\arraybackslash}m{1.25cm}|}{\textbf{Te. MSE}}
& \multicolumn{1}{>{\centering\arraybackslash}m{1.25cm}|}{\textbf{Te. RMSE}}
\\
\hline
\hline
LR
& 0.9129706418 & 0.2504014163 & 0.5004012553
& 0.9130137709 & 0.2507766686 & 0.5007760663 \\
\hline
RF
& 0.9870627444 & 0.03722315308 & 0.1929330275
& 0.9077351184 & 0.2659947428 & 0.5157467817 \\
\hline
DT
& 0.583743874773 & 1.504864234 & 1.226729487
& 0.8151677147 & 0.5328616407 & 0.7299737261 \\
\hline
Ridge Regression
& 0.9129706417 & 0.2504014163 & 0.5004012553
& 0.9130137621 & 0.250776694 & 0.5007760917 \\
\hline
Lasso Regression
& 0.8847181529 & 0.3304832303 & 0.5748767088
& 0.885137457 & 0.3323514293 & 0.5764992882 \\
\hline
\end{tabular}%
}
\end{table}

\begin{table}[H]
\centering
\caption{Regression performance for the 80:20 train--test split using RandomizedSearchCV with 24 features.}
\label{tab:results-8020-random-24}
\scriptsize
\setlength{\tabcolsep}{3pt}
\renewcommand{\arraystretch}{1.5}

\resizebox{\textwidth}{!}{%
\begin{tabular}{|
>{\centering\arraybackslash}m{1.7cm}|
r|r|r|r|r|r|}
\hline
\hline
\textbf{Model}
& \multicolumn{1}{>{\centering\arraybackslash}m{1.25cm}|}{\textbf{Tr. $R^2$}}
& \multicolumn{1}{>{\centering\arraybackslash}m{1.25cm}|}{\textbf{Tr. MSE}}
& \multicolumn{1}{>{\centering\arraybackslash}m{1.25cm}|}{\textbf{Tr. RMSE}}
& \multicolumn{1}{>{\centering\arraybackslash}m{1.25cm}|}{\textbf{Te. $R^2$}}
& \multicolumn{1}{>{\centering\arraybackslash}m{1.25cm}|}{\textbf{Te. MSE}}
& \multicolumn{1}{>{\centering\arraybackslash}m{1.25cm}|}{\textbf{Te. RMSE}}
\\
\hline
\hline
LR
& 0.9129705515 & 0.2504016758 & 0.5004015146
& 0.9130136856 & 0.2507769147 & 0.500776312 \\
\hline
RF
& 0.941970508 & 0.1669628189 & 0.4086108404
& 0.9064520849 & 0.26969366 & 0.5193203828 \\
\hline
DT
& 0.9161022839 & 0.2413910359 & 0.4913156174
& 0.909994562 & 0.2594808871 & 0.5093926649 \\
\hline
Ridge Regression
& 0.9129706415 & 0.250401417 & 0.500401256
& 0.9130137195 & 0.2507768168 & 0.5007762143 \\
\hline
Lasso Regression
& 0.5846677618 & 1.194996525 & 1.093158966
& 0.5836824697 & 1.200221281 & 1.095546111 \\
\hline
\end{tabular}%
}
\end{table}

\begin{table}[H]
\centering
\caption{Regression performance for the 80:20 train--test split using GridSearchCV with 24 features.}
\label{tab:results-8020-grid-24}
\scriptsize
\setlength{\tabcolsep}{3pt}
\renewcommand{\arraystretch}{1.5}

\resizebox{\textwidth}{!}{%
\begin{tabular}{|
>{\centering\arraybackslash}m{1.7cm}|
r|r|r|r|r|r|}
\hline
\hline
\textbf{Model}
& \multicolumn{1}{>{\centering\arraybackslash}m{1.25cm}|}{\textbf{Tr. $R^2$}}
& \multicolumn{1}{>{\centering\arraybackslash}m{1.25cm}|}{\textbf{Tr. MSE}}
& \multicolumn{1}{>{\centering\arraybackslash}m{1.25cm}|}{\textbf{Tr. RMSE}}
& \multicolumn{1}{>{\centering\arraybackslash}m{1.25cm}|}{\textbf{Te. $R^2$}}
& \multicolumn{1}{>{\centering\arraybackslash}m{1.25cm}|}{\textbf{Te. MSE}}
& \multicolumn{1}{>{\centering\arraybackslash}m{1.25cm}|}{\textbf{Te. RMSE}}
\\
\hline
\hline
LR
& 0.9129705515 & 0.2504016758 & 0.5004015146
& 0.9130136856 & 0.2507769147 & 0.500776312 \\
\hline
RF
& 0.941970508 & 0.1669628189 & 0.4086108404
& 0.9064520849 & 0.26969366 & 0.5193203828 \\
\hline
DT
& 0.9134186857 & 0.2491123014 & 0.499111512
& 0.9120196948 & 0.2536425372 & 0.5036293649 \\
\hline
Ridge Regression
& 0.9129706417 & 0.2504014163 & 0.5004012553
& 0.9130137621 & 0.250776694 & 0.5007760917 \\
\hline
Lasso Regression
& 0.9129703752 & 0.2504021831 & 0.5004020215
& 0.9130138109 & 0.2507765533 & 0.5007759512 \\
\hline
\end{tabular}%
}
\end{table}

 \begin{table}[H]
\centering
\caption{Regression performance for the 70:30 train--test split without hyperparameter tuning using 20 features.}
\label{tab:results-7030-untuned-20}
\scriptsize
\setlength{\tabcolsep}{3pt}
\renewcommand{\arraystretch}{1.5}

\resizebox{\textwidth}{!}{%
\begin{tabular}{|
>{\centering\arraybackslash}m{1.7cm}|
r|r|r|r|r|r|}
\hline
\hline
\textbf{Model}
& \multicolumn{1}{>{\centering\arraybackslash}m{1.25cm}|}{\textbf{Tr. $R^2$}}
& \multicolumn{1}{>{\centering\arraybackslash}m{1.25cm}|}{\textbf{Tr. MSE}}
& \multicolumn{1}{>{\centering\arraybackslash}m{1.25cm}|}{\textbf{Tr. RMSE}}
& \multicolumn{1}{>{\centering\arraybackslash}m{1.25cm}|}{\textbf{Te. $R^2$}}
& \multicolumn{1}{>{\centering\arraybackslash}m{1.25cm}|}{\textbf{Te. MSE}}
& \multicolumn{1}{>{\centering\arraybackslash}m{1.25cm}|}{\textbf{Te. RMSE}}
\\
\hline
\hline
LR
& 0.912930456 & 0.2503510489 & 0.5003509258
& 0.9130916683 & 0.2507726187 & 0.5007720227 \\
\hline
RF
& 0.9870013283 & 0.03737507915 & 0.193326354
& 0.9074236222 & 0.2671276761 & 0.5168439572 \\
\hline
DT
& 1 & 0 & 0
& 0.8159752718 & 0.531000447 & 0.7286977748 \\
\hline
Ridge Regression
& 0.912930456 & 0.250351049 & 0.5003509258
& 0.9130916633 & 0.250772633 & 0.500772037 \\
\hline
Lasso Regression
& 0.585175823 & 1.192743905 & 1.092128154
& 0.5858101289 & 1.195138332 & 1.093223825 \\
\hline
\end{tabular}%
}
\end{table}

\begin{table}[H]
\centering
\caption{Regression performance for the 70:30 train--test split using RandomizedSearchCV with 20 features.}
\label{tab:results-7030-random-20}
\scriptsize
\setlength{\tabcolsep}{3pt}
\renewcommand{\arraystretch}{1.5}

\resizebox{\textwidth}{!}{%
\begin{tabular}{|
>{\centering\arraybackslash}m{1.7cm}|
r|r|r|r|r|r|}
\hline
\hline
\textbf{Model}
& \multicolumn{1}{>{\centering\arraybackslash}m{1.25cm}|}{\textbf{Tr. $R^2$}}
& \multicolumn{1}{>{\centering\arraybackslash}m{1.25cm}|}{\textbf{Tr. MSE}}
& \multicolumn{1}{>{\centering\arraybackslash}m{1.25cm}|}{\textbf{Tr. RMSE}}
& \multicolumn{1}{>{\centering\arraybackslash}m{1.25cm}|}{\textbf{Te. $R^2$}}
& \multicolumn{1}{>{\centering\arraybackslash}m{1.25cm}|}{\textbf{Te. MSE}}
& \multicolumn{1}{>{\centering\arraybackslash}m{1.25cm}|}{\textbf{Te. RMSE}}
\\
\hline
\hline
LR
& 0.9129301202 & 0.2503520145 & 0.5003518907
& 0.9130924958 & 0.2507702309 & 0.5007696385 \\
\hline
RF
& 0.9440043794 & 0.1610042011 & 0.4012532879
& 0.9094332743 & 0.2613288568 & 0.5112033419 \\
\hline
DT
& 0.9159929172 & 0.2415455547 & 0.4914728422
& 0.9101203013 & 0.2593464514 & 0.509260691 \\
\hline
Ridge Regression
& 0.9129304557 & 0.25035105 & 0.5003509268
& 0.9130916393 & 0.2507727024 & 0.5007721063 \\
\hline
Lasso Regression
& 0.5842710734 & 1.195345331 & 1.093318495
& 0.584907467 & 1.195345331 & 1.09331818164 \\
\hline
\end{tabular}%
}
\end{table}

\begin{table}[H]
\centering
\caption{Regression performance for the 70:30 train--test split using GridSearchCV with 20 features.}
\label{tab:results-7030-grid-20}
\scriptsize
\setlength{\tabcolsep}{3pt}
\renewcommand{\arraystretch}{1.5}

\resizebox{\textwidth}{!}{%
\begin{tabular}{|
>{\centering\arraybackslash}m{1.7cm}|
r|r|r|r|r|r|}
\hline
\hline
\textbf{Model}
& \multicolumn{1}{>{\centering\arraybackslash}m{1.25cm}|}{\textbf{Tr. $R^2$}}
& \multicolumn{1}{>{\centering\arraybackslash}m{1.25cm}|}{\textbf{Tr. MSE}}
& \multicolumn{1}{>{\centering\arraybackslash}m{1.25cm}|}{\textbf{Tr. RMSE}}
& \multicolumn{1}{>{\centering\arraybackslash}m{1.25cm}|}{\textbf{Te. $R^2$}}
& \multicolumn{1}{>{\centering\arraybackslash}m{1.25cm}|}{\textbf{Te. MSE}}
& \multicolumn{1}{>{\centering\arraybackslash}m{1.25cm}|}{\textbf{Te. RMSE}}
\\
\hline
\hline
LR
& 0.9129301202 & 0.2503520145 & 0.5003518907
& 0.9130924958 & 0.2507702309 & 0.5007696385 \\
\hline
RF
& 0.9440043794 & 0.1610042011 & 0.4012532879
& 0.9094332743 & 0.2613288568 & 0.5112033419 \\
\hline
DT
& 0.913482506 & 0.2487637397 & 0.4987622076
& 0.9119783059 & 0.2539852084 & 0.5039694518 \\
\hline
Ridge Regression
& 0.912930456 & 0.250351049 & 0.5003509258
& 0.9130916633 & 0.250772633 & 0.500772037 \\
\hline
Lasso Regression
& 0.9129301542 & 0.2503519168 & 0.5003517936   
& 0.9130926394 & 0.2507698165 & 0.5007692248 \\
\hline
\end{tabular}%
}
\end{table}

\begin{table}[H]
\centering
\caption{Regression performance for the 70:30 train--test split without hyperparameter tuning using 24 features.}
\label{tab:results-7030-untuned-24}
\scriptsize
\setlength{\tabcolsep}{3pt}
\renewcommand{\arraystretch}{1.5}

\resizebox{\textwidth}{!}{%
\begin{tabular}{|
>{\centering\arraybackslash}m{1.7cm}|
r|r|r|r|r|r|}
\hline
\hline
\textbf{Model}
& \multicolumn{1}{>{\centering\arraybackslash}m{1.25cm}|}{\textbf{Tr. $R^2$}}
& \multicolumn{1}{>{\centering\arraybackslash}m{1.25cm}|}{\textbf{Tr. MSE}}
& \multicolumn{1}{>{\centering\arraybackslash}m{1.25cm}|}{\textbf{Tr. RMSE}}
& \multicolumn{1}{>{\centering\arraybackslash}m{1.25cm}|}{\textbf{Te. $R^2$}}
& \multicolumn{1}{>{\centering\arraybackslash}m{1.25cm}|}{\textbf{Te. MSE}}
& \multicolumn{1}{>{\centering\arraybackslash}m{1.25cm}|}{\textbf{Te. RMSE}}
\\
\hline
\hline
LR
& 0.912930456 & 0.2503510489 & 0.5003509258
& 0.9130916683 & 0.2507726187 & 0.5007720227 \\
\hline
RF
& 0.9870488888 & 0.03723832849 & 0.1929723516
& 0.9077328853 & 0.2662353024 & 0.5159799438 \\
\hline
DT
& 0.58374388383 & 1.504864284 & 1.2267282564
& 0.8157054124 & 0.5317791221 & 0.7292318713 \\
\hline
Ridge Regression
& 0.912930456 & 0.250351049 & 0.5003509258
& 0.9130916634 & 0.250772633 & 0.500772036 \\
\hline
Lasso Regression
& 0.585175823 & 0.585776991 &  0.7653606934
& 0.5858101289 & 1.195138332 & 1.093223366 \\
\hline
\end{tabular}%
}
\end{table}

\begin{table}[H]
\centering
\caption{Regression performance for the 70:30 train--test split using RandomizedSearchCV with 24 features.}
\label{tab:results-7030-random-24}
\scriptsize
\setlength{\tabcolsep}{3pt}
\renewcommand{\arraystretch}{1.5}

\resizebox{\textwidth}{!}{%
\begin{tabular}{|
>{\centering\arraybackslash}m{1.7cm}|
r|r|r|r|r|r|}
\hline
\hline
\textbf{Model}
& \multicolumn{1}{>{\centering\arraybackslash}m{1.25cm}|}{\textbf{Tr. $R^2$}}
& \multicolumn{1}{>{\centering\arraybackslash}m{1.25cm}|}{\textbf{Tr. MSE}}
& \multicolumn{1}{>{\centering\arraybackslash}m{1.25cm}|}{\textbf{Tr. RMSE}}
& \multicolumn{1}{>{\centering\arraybackslash}m{1.25cm}|}{\textbf{Te. $R^2$}}
& \multicolumn{1}{>{\centering\arraybackslash}m{1.25cm}|}{\textbf{Te. MSE}}
& \multicolumn{1}{>{\centering\arraybackslash}m{1.25cm}|}{\textbf{Te. RMSE}}
\\
\hline
\hline
LR
& 0.912930186 & 0.2503518254 & 0.5003517018
& 0.9130922801 & 0.2507708534 & 0.5007702601 \\
\hline
RF
& 0.9416447868 & 0.1677887372 & 0.4096202354
& 0.906353173 & 0.2702164405 & 0.5198234705 \\
\hline
DT
& 0.9160351483 & 0.2414241278 & 0.4913492931
& 0.9100678899 & 0.2594976836 & 0.5094091515 \\
\hline
Ridge Regression
& 0.9129304557 & 0.25035105 & 0.5003509268
& 0.9130916393 & 0.2507727024 & 0.5007721063 \\
\hline
Lasso Regression
& 0.5842710734 & 1.195345331 & 1.093318495
& 0.584907467 & 1.197742949 & 1.094414432 \\
\hline
\end{tabular}%
}
\end{table}

\begin{table}[H]
\centering
\caption{Regression performance for the 70:30 train--test split using GridSearchCV with 24 features.}
\label{tab:results-7030-grid-24}
\scriptsize
\setlength{\tabcolsep}{3pt}
\renewcommand{\arraystretch}{1.5}

\resizebox{\textwidth}{!}{%
\begin{tabular}{|
>{\centering\arraybackslash}m{1.7cm}|
r|r|r|r|r|r|}
\hline
\hline
\textbf{Model}
& \multicolumn{1}{>{\centering\arraybackslash}m{1.25cm}|}{\textbf{Tr. $R^2$}}
& \multicolumn{1}{>{\centering\arraybackslash}m{1.25cm}|}{\textbf{Tr. MSE}}
& \multicolumn{1}{>{\centering\arraybackslash}m{1.25cm}|}{\textbf{Tr. RMSE}}
& \multicolumn{1}{>{\centering\arraybackslash}m{1.25cm}|}{\textbf{Te. $R^2$}}
& \multicolumn{1}{>{\centering\arraybackslash}m{1.25cm}|}{\textbf{Te. MSE}}
& \multicolumn{1}{>{\centering\arraybackslash}m{1.25cm}|}{\textbf{Te. RMSE}}
\\
\hline
\hline
LR
& 0.912930186 & 0.2503518254 & 0.5003517018
& 0.9130922801 & 0.2507708534 & 0.5007702601 \\
\hline
RF
& 0.9416447868 & 0.1677887372 & 0.4096202354
& 0.906353173 & 0.2702164405 & 0.5198234705 \\
\hline
DT
& 0.913483205 & 0.2487617301 & 0.4987601929
& 0.9119788539 & 0.253983627 & 0.5039678829 \\
\hline
Ridge Regression
& 0.912930456 & 0.250351049 & 0.5003509258
& 0.9130916634 & 0.250772633 & 0.5007720369 \\
\hline
Lasso Regression
& 0.912930177 & 0.2503518512 & 0.5003517275
& 0.9130925912 & 0.2507699556 & 0.5007693637 \\
\hline
\end{tabular}%
}
\end{table}

\end{document}